\documentclass{article}

\usepackage{PRIMEarxiv}

\usepackage[utf8]{inputenc} 
\usepackage[T1]{fontenc}    
\usepackage{hyperref}       
\usepackage{url}            
\usepackage{booktabs}       
\usepackage{amsfonts}       
\usepackage{nicefrac}       
\usepackage{microtype}      
\usepackage{lipsum}
\usepackage{fancyhdr}       
\usepackage{graphicx}       
\graphicspath{{media/}}     

\usepackage{natbib}
\usepackage{algorithm}
\usepackage{algorithmic}

\usepackage{amsfonts}       
\usepackage{amsmath}
\usepackage{amsthm}

\usepackage{booktabs}
\usepackage{multirow}

\usepackage{color}
\usepackage[none]{hyphenat}

\newtheorem{thm}{Theorem}
\newtheorem{assump}{Assumption}

\newtheorem{lem}{Lemma}

\title{Robust Tests in Online Decision-Making
\thanks{Codes for experiments can be found at \texttt{https://github.com/gisoo1989/RobustTests}.} 
}

\author{
  Gi-Soo Kim \\
  Department of Industrial Engineering \&  \\
  Artificial Intelligence Graduate School \\
  UNIST\\
  \texttt{gisookim@unist.ac.kr} \\
   \And
  Hyun-Joon Yang \\
  Department of Psychiatry and Behavioral Sciences \\
  Stanford University School of Medicine \\
  \texttt{yanghyun@stanford.edu} \\
     \AND
  Jane P. Kim \\
  Department of Psychiatry and Behavioral Sciences \\
  Stanford University School of Medicine \\
  \texttt{janepkim@stanford.edu} \\
}
\begin{document}
\maketitle

\begin{abstract}
Bandit algorithms are widely used in sequential decision problems to maximize the cumulative reward. One potential application is mobile health, where the goal is to promote the user's health through personalized interventions based on user specific information acquired through wearable devices. Important considerations include the type of, and frequency with which data is collected (e.g. GPS, or continuous monitoring), as such factors can severely impact app performance and users’ adherence. In order to balance the need to collect data that is useful with the constraint of impacting app performance, one needs to be able to assess the usefulness of variables. Bandit feedback data are sequentially correlated, so traditional testing procedures developed for independent data cannot apply. Recently, a statistical testing procedure was developed for the actor-critic bandit algorithm. An actor-critic algorithm maintains two separate models, one for the actor, the action selection policy, and the other for the critic, the reward model. The performance of the algorithm as well as the validity of the test are guaranteed only when the critic model is correctly specified. However, misspecification is frequent in practice due to incorrect functional form or missing covariates. In this work, we propose a modified actor-critic algorithm which is robust to critic misspecification and derive a novel testing procedure for the actor parameters in this case. 
\end{abstract}


\section{Introduction}

Bandit algorithms apply to sequential decision problems. We assume a set of candidate actions, or {\it arms}, is revealed sequentially to a learning agent along with side information called contexts. The agent can pull one arm at a time and receives a corresponding reward. 
The expected value of the reward is an unknown function of the context information of the chosen action. The goal of the agent is to adaptively learn an action allocation policy so as to achieve high cumulative reward. The main challenge is the exploration-exploitation trade-off, which represents the dilemma between pulling arms that the agent is uncertain about for the sake of learning (exploration) and pulling the best arm based on current, limited knowledge (exploitation).

Bandit algorithms can be particularly useful in the context of personalizing health interventions in Mobile Health \citep{Tewari17}. The goal of Mobile Health (mHealth) apps is to promote the user's health through personalized interventions tailored to the user specific information acquired through devices such as phones or wearable devices. One important issue related to mHealth apps is that the frequency of data queries (e.g. queries to the Health Kit API) impacts the app performance. As queries become increasingly frequent, the processing time slows down the app from the user perspective, which can result in low adherence to the app and hence low reward. Hence, the frequency of data queries and the reward feedback go hand in hand. Beyond performance related costs, there could also be costs of ethical valence (e.g. privacy) associated with querying data and thus it is important to collect only data that are correlated with the reward. Currently, there is little work on assessing the utility of variables collected by wearables.

We consider testing the utility of context variables for an actor-critic bandit algorithm \citep{Lei17}. An actor-critic bandit algorithm maintains two separate parameterized models, one for the actor, the action allocation policy, and the other for the critic, the reward model. {The focus of this work is on testing the  variables used in the actor model, which requires that asymptotic distributions of the actor parameter estimates are known.}   \citet{Lei17} proved that when the reward model is linear and is correctly specified by the critic, then the actor parameter estimates converge in probability to the parameters of the optimal policy and asymptotically follow a normal distribution.

Based on the asymptotic normality of the actor parameter estimates, we can apply a Z-test to assess the significance of the actor parameters. The validity of model-based tests, such as those of  \citet{Lei17}, relies on the assumption that the linear model is correctly specified; in other words, that the assumed statistical model represents the true reward function.  Linear functions may, however, fail to accurately represent the true nature of the reward function, and often there is no a priori reason to hypothesize the reward should be of a certain functional form.  When the parameterized critic model is not correctly specified (i.e. the true reward model is of a different form than the working model),  asymptotic normality may not hold. In this paper, we propose a new actor-critic algorithm and a testing procedure that is robust to the misspecification of the critic model. The main contributions of our paper are as follows:
\begin{itemize}
    \item We propose a new actor-critic algorithm where the actor parameter {estimates} converge to the parameters of the optimal policy even when the reward model is misspecified by the critic.
    \item We show that in the new algorithm, critic and actor parameter estimates asymptotically follow a normal distribution.
\item We conduct experiments on synthetic data and {real data} and show that our testing procedure appropriately assess the significance of the parameters.
\end{itemize}

\subsection{Related Works on Robust Bandits}

{Our work is distinct from the limited literature on robust bandits. First, the focus of the existing works \citep{Tang21, Hao19, Zhu18} {is on the robustness to the noise of the rewards.} For example, \citet{Tang21} and \citet{Hao19} developed Upper-Confidence Bound (UCB) algorithms without requiring to specify the tail property of the reward, and \citet{Zhu18} developed an actor-critic (AC) algorithm that is robust to outliers. These methods however, are built upon the linearity of the reward model. \citet{Ghosh17} developed a new algorithm that maintains the sublinear regret in a model with large deviations from linearity, but is restricted to the case where the action feature set is fixed over time. Hence even under large deviations, the problem can still be addressed by a multi-armed bandit algorithm with regret scaling with the number of arms instead of feature dimension.}

{Second, we consider the AC algorithm, which, apart from $\epsilon$-greedy, is unique in that asymptotic distributions of the parameter estimates are known; there are currently no established statistical testing procedures for the UCB or Thompson sampling algorithms.
We consider the impact of misspecification on the validity of inferential testing of the utility of contextual variables (i.e. significance of actor parameters in the AC algorithm). To the best of our knowledge, no other work addresses the {robustness of} inferential testing in the context of the actor-critic algorithm. {In a related work on the $\epsilon$-greedy bandit,  Chen, Lu, and Song (2020) derived the asymptotic properties of a weighted least squares estimator (LSE) for misspecified reward models. The authors demonstrated that the weighted LSE has asymptotic normal distribution with the mean being the least false linear parameter. While both this and our approach offer robust tests, ours offers a directed approach to exploration, which may be efficient and desirable in the mobile health setting where the action space is large.}}

\section{Preliminaries}

\subsection{Problem Formulation}

We first formulate the bandit problem. At each time \(t\), the learning agent can pull one arm among \(N\) alternative arms. The \(i\)-th arm (\(i=1,\cdots,N\)) returns a random reward \(r_{t,i}\) with unknown mean when it is pulled. Prior to arm selection, a finite-dimensional context vector \(b_{t,i}\in \mathbb{R}^d\) for each arm \(i\) is revealed to the agent. The agent tailors his(her) choice based on this contextual information. Let $a_t$ denote the arm index pulled at time $t$. Then the goal of the agent is to maximize the cumulative sum of the rewards $r_{t,a_t}$ over a finite time horizon $T$. {We assume that the full set of contexts $b_t=\{b_{t,1}, \cdots, b_{t,N}\}$ {and the full set of rewards $r_t=\{r_{t,1},\cdots, r_{t,N}\}$} are independently and identically (i.i.d.) distributed over time.} 
{Also, without loss of generality, we assume the $L_2$-norm of $b_{t,i}$ is bounded by 1, i.e., $||b_{t,i}||_2\leq 1$. }

\subsection{Notations}

We denote the $L_2$-norm of a vector $x$ as $||x||_2$, the set of natural numbers from 1 to $N$ as $[N]$, the set of all natural numbers as $\mathbb{N}$, the $d$-dimensional identity matrix as $I^{d\times d}$, and the $d$-dimensional vector with all elements equal to $0$ as $0_d$.

\subsection{Actor-Critic Bandit Algorithm {for Linear Rewards}}

Under linear reward assumption $\mathbb{E}[r_{t,i}|b_{t,i}]=b_{t,i}^T\mu^*$ for some $\mu^*\in\mathbb{R}^d$ {with $||\mu^*||_2\leq 1$}, \citet{Lei17} proposed the actor-critic bandit algorithm (Algorithm 1) which learns two parametrized models, the critic and the actor. The critic for the $i$-th arm is a linear function of the $i$-th context variable with parameter $\mu\in\mathbb{R}^d$, $b_{t,i}^T\mu$. The actor is the action allocation probability and is parametrized by a {softmax function} with parameter $\theta\in\mathbb{R}^d$, i.e., the probability of pulling the $i$-th arm at time $t$ is
$\pi_{\theta}(b_t,i)={\mathrm{exp}(b_{t,i}^T\theta)}/\{\sum_{j=1}^N\mathrm{exp}(b_{t,j}^T\theta)\}.$  \citet{Lei17} define the optimal parameter $\theta^*$ as the value that maximizes the penalized expected reward, 
$$\theta^*=\underset{\theta}{\mathrm{argmax}}~\mathbb{E}\left[\sum_{i=1}^Nb_{t,i}^T\mu~\pi_{\theta}(b_{t},i)\right]-\lambda \theta^T\theta,$$
where $\lambda>0$ and the expectation is taken over the distribution of $b_t$. 
The penalty term $-\lambda\theta^T\theta$\footnote{The presented penalty form is a special case of the penalty  proposed in \citet{Lei17}.} is introduced to constrain the norm of $\theta$. Due to the penalty, there exists $\gamma>0$ such that $\gamma <{\pi}_{{\theta}^*}(b_t,i)<1-\gamma$ for every $i$ with high probability. This guarantees treatment variety, which prevents habituation and increases user engagement in many applications including mHealth. Also, when the expected rewards of the arms are the same so that the $\mathbb{E}\left[\sum_{i=1}^Nb_{t,i}^T\mu~\pi_{\theta}(b_{t},i)\right]$ term does not change according to the values of $\theta$, $\theta^*$ is unique at $\theta^*=0_d$.

\begin{algorithm}[tb]
\label{alg1}
\caption{Actor-Critic algorithm for linear reward [Lei et al.,2017]}
\begin{algorithmic}[1]
\STATE Set $B=\xi I^{d\times d}$, $y=0_d$, $\lambda>0$, $\xi>0$.
\FOR{$t=1,\cdots, T$}
\STATE {Pull arm $a_t$ according to probability $\left\{{\pi}_{\hat{\theta}_{t-1}}(b_t,i)\right\}_{i=1}^N$}  and get reward $r_{t,a_t}.$
\STATE {\bf Critic update}: $$B\leftarrow B+b_{t,a_t}b_{t,a_t}^T,~~~y\leftarrow y+b_{t,a_t}r_{t,a_t},~~\hat{\mu}_t\leftarrow B^{-1}y.$$
\STATE $\hat{r}_{\tau,i}\leftarrow\mathrm{max}(-2,\mathrm{min}(2, b_{\tau,i}^T\hat{\mu}_t))$ for $i\in[N]$, $\tau\in [t].$
\STATE {\bf Actor update}:  
$$\hat{\theta}_t\leftarrow \underset{\theta}{\mathrm{argmax}}\frac{1}{t}\sum_{\tau=1}^t\sum_{i=1}^N\hat{r}_{\tau,i}\pi_{\theta}(b_{\tau},i)-\lambda \theta^T\theta.$$
\ENDFOR
\end{algorithmic}
\end{algorithm}

The estimator $\hat{\mu}$ of the critic parameter $\mu$ is the Ridge estimator using the context and reward pair of the chosen arms. The estimator $\hat{\theta}$ of the policy parameter $\theta$ is the maximizer of the estimate of the penalized expected reward, $\frac{1}{t}\sum_{\tau=1}^t\sum_{i=1}^N\hat{r}_{\tau,i}\pi_{\theta}(b_{\tau},i)-\lambda \theta^T\theta$, where $\hat{r}_{\tau,i}$ is the truncated estimate of the reward defined in Algorithm 1, line 5. When the true critic parameter $\mu^*$ has $||\mu^*||_2\leq 1$ and $\hat{\mu}_t$ converges to $\mu^*$, the truncated reward estimate $\hat{r}_{t,i}$ approaches the untruncated estimate, $b_{t,i}^T\hat{\mu}_t.$ The boundedness of $\hat{r}_{\tau,i}$ and the penalty term ensures that $\hat{\theta}_t$ is bounded. This guarantees that there exists $\gamma>0$ such that $\gamma <{\pi}_{\hat{\theta}_{t}}(b_t,i)<1-\gamma$ for every $i$. This prevents $ {\pi}_{\hat{\theta}_{t}}(b_t,i)$ from concentrating on a single arm and induces a degree of exploration.  

\citet{Lei17} showed that under some regular assumptions on the distribution of the contexts and rewards, $\hat{\mu}_t$ and $\hat{\theta}_t$ converge in probability to $\mu^*$ and $\theta^*$ respectively and are asymptotically normally distributed, hereby enabling a testing procedure.  

\subsection{Misspecification of Models}
The validity of model-based testing is predicated on correctly specified models.  However, misspecification is frequent in practice due to incorrect functional forms or missing covariates. In the statistics literature, robustness has been considered in the context of using models to test causal effects from data collected in experiments. Linear and GLM regression models \citep{RV} and proportional and multiplicative hazards models \citep{Kim13} have been shown to be robust to misspecification when considering the test of the coefficient of the treatment assignment in the context of randomized trials. However in bandit settings, asymptotic normality is not guaranteed to hold when the working model is incorrect. In this paper, we consider the case where the critic is misspecified.

\subsection{Inference from Bandit Feedback Data}

Besides \citet{Lei17}, there is a recent growing body of literature on deriving the distribution of the parameter estimates from bandit feedback data. Bandit feedback data are not i.i.d. but are correlated due to adaptivity. This causes complexity in deriving the distribution of the estimates. \citet{Zhang21} recently showed the asymptotic distribution of M-estimators from bandit data. This work considered correctly specified reward models only.
\citet{Chen20} derived the asymptotic normality of the ordinary and weighted least-squares estimators when data are accumulated by a $\varepsilon$-greedy algorithm, in both cases where the reward model is linear or not linear. When the reward model is not linear, they proved that the estimator with inverse-probability weighting converges to the normal distribution with mean being the least false parameter in terms of the population distribution of the contexts. {Since the action decision in $\varepsilon$-greedy algorithms is based on reward estimate values, a robust test on the utility of the variables could be conducted by testing the significance of the least false parameters. However, as aforementioned earlier, we note that $\varepsilon$-greedy performs uniform exploration over context spaces which may be undesirable when the action space is large.  }

\subsection{Compatibility Condition in Actor-Critic Algorithm}

The algorithm of \citet{Lei17} and the theoretical derivation therein exploit the fact that the true reward model is linear. The true nature of the reward can however be far from linear.  \citet{Sutton99} proved the following Lemma \ref{suttonlem} which implies that if the reward model and policy model {are both differentiable with respect to their parameters and }satisfy the compatibility condition, the algorithm converges to the optimal policy $\pi_{\theta*}$ though the critic model may be misspecified. If we denote the critic model parameterized by $\mu$ as $m_{\mu}(\cdot),$ the compatibility condition states:
\begin{equation}{\dot{\pi}_{\theta}(b_t,i)}/{\pi_{\theta}(b_t,i)}=\dot{m}_{\mu}(b_{t,i}),\label{comp}\end{equation}
where $\dot{\pi}_{\theta}(b_t,i)=\frac{\partial}{\partial\theta}\pi_{\theta}(b_t,i)$ and $\dot{m}_{\mu}(b_{t,i})=\frac{\partial}{\partial \mu}m_{\mu}(b_{t,i})$.
\begin{lem}\label{suttonlem} (Theorem 2 of \citet{Sutton99})
Let \begin{equation}J(\theta)=\mathbb{E}_{b,r}\left[\sum_{i=1}^Nr_{t,i}\pi_{\theta}(b_t,i)\right]-\lambda \theta^T\theta\label{obj},\end{equation} where $\mathbb{E}_{b,r}(\cdot)$ denotes the expectation over both the context and reward. Suppose the critic parameter $\mu$ minimizes 
\begin{align*}U(\mu,\theta):=\mathbb{E}_{b,r}\left[\sum_{i=1}^N\left\{r_{t,i}-m_{\mu}(b_{t,i})\right\}^2\pi_{\theta}(b_t,i)\right],\end{align*}
and the actor parameter $\theta$ maximizes
\begin{align*}J(\mu, \theta):=\mathbb{E}_{b}\left[\sum_{i=1}^Nm_{\mu}(b_{t,i})~\pi_{\theta}(b_t,i)\right]-\lambda \theta^T\theta.\end{align*}
Then if $\pi_{\theta}(\cdot)$ and $m_{\mu}(\cdot)$ satisfy the compatibility condition (\ref{comp}), the actor parameter $\theta$ satisfies $\frac{\partial}{\partial \theta}J(\theta)=0$.
\end{lem}
\begin{proof} The parameters $\mu$ and $\theta$ jointly solve $U_{\mu}(\mu,\theta)=0$ and $J_{\theta}(\mu,\theta)=0$, where $U_{\mu}(\mu,\theta)=-\frac{1}{2}\frac{\partial}{\partial \mu}U(\mu,\theta)$ and $J_{\theta}(\mu,\theta)=\frac{\partial}{\partial \theta}J(\mu,\theta)$. We have,
\begin{align}U_{\mu}(\mu,\theta)=\mathbb{E}_{b,r}\left[\sum_{i=1}^N\left\{r_{t,i}-m_{\mu}(b_{t,i})\right\}\dot{m}_{\mu}(b_{t,i})\pi_{\theta}(b_t,i)\right]\label{comp1}\end{align}
and 
\begin{align}J_{\theta}(\mu,\theta)=\mathbb{E}_b\left[\sum_{i=1}^Nm_{\mu}(b_{t,i})~\dot{\pi}_{\theta}(b_t,i)\right]-2\lambda \theta\label{comp2}\end{align}
Due to (\ref{comp}) and the facts that (\ref{comp1})$=0$, and (\ref{comp2})$=0$, the parameter $\theta$ satisfies
$$\frac{\partial}{\partial\theta}J(\theta)=\mathbb{E}_{b,r}\left[\sum_{i=1}^Nr_{t,i}~\dot{\pi}_{\theta}(b_t,i)\right]-2\lambda \theta=0.$$
\end{proof}
Note that in Lemma \ref{suttonlem}, $J(\theta)$ is defined in terms of the true rewards $r_{t,i}$'s, while $J(\mu,\theta)$ replaces them with $m_{\mu}(b_t,i)$'s. If the true reward model is linear, i.e., if $\mathbb{E}[r_{t,i}|b_{t,i}]=b_{t,i}^T\mu^*$, and if $m_{\mu}(b_t,i)=b_{t,i}^T\mu$, then we have $J(\theta)=J(\mu^*,\theta)$. However when the true model is not linear, $J(\theta)$ and $J(\mu,\theta)$ are completely different functions. 
\citet{Sutton99} show that (\ref{comp}) is satisfied when the actor model is a {softmax function} and the critic model is linear in the same context vectors as the policy, except they should be centered to have { weighted }mean 0:
\begin{align}
    \text{\bf critic}:~~m_{\mu,\theta}(b_{t,i})&= \left\{b_{t,i}-\sum_{j=1}^N\pi_{\theta}(b_t,j)b_{t,j}\right\}^T\mu\label{critic}\\
    \text{\bf actor}:~~~~~\pi_{\theta}(b_t,i)&=\frac{\mathrm{exp}\left(b_{t,i}^T\theta\right)}{\sum_{j=1}^N\mathrm{exp}\left(b_{t,j}^T\theta\right)}\label{actor}
\end{align}
 The model (\ref{critic}) can be viewed as the approximation of the {\it advantage} function \citep{Baird93}. {The advantage function enables to discard variables that do not vary by arm (e.g., age of the user).
 We would still need such variables if we model the reward instead of the advantage.} From now on we denote the model (\ref{critic}) as $m_{\mu,\theta}(\cdot)$ instead of $m_{\mu}(\cdot)$ to show its dependency on $\theta$ as well. 
 Meanwhile, since $\sum_{i=1}^N\dot{\pi}_{\theta}(b_t,i)=0$, equation (\ref{comp2}) is equivalent to 
$$J_{\theta}(\mu,\theta)=\mathbb{E}_b\left[\sum_{i=1}^Nb_{t,i}^T\mu~\dot{\pi}_{\theta}(b_t,i)\right]-2\lambda \theta.$$
So we redefine \begin{align}J(\mu, \theta)=\mathbb{E}_{b}\left[\sum_{i=1}^N{b_{t,i}^T\mu}~\pi_{\theta}(b_t,i)\right]-\lambda \theta^T\theta.\label{obj_theta}\end{align}
We can find the value of $\theta$ satisfying $(\ref{comp2})=0$ as the maximizer of the redefined $J(\mu,\theta)$.

\section{Proposed Algorithm}

We propose a new actor-critic algorithm which uses (\ref{critic}) and (\ref{actor}) to model the reward and action selection policy. We consider the case where the true functional form of the reward model may not be linear. In this case, we re-define the target parameters $\mu^*$ and $\theta^*$ as  
    $$\theta^*=\underset{\theta}{\mathrm{argmax}}~J(\theta),~~\mu^*=\underset{\mu}{\mathrm{argmin}}~U(\mu,\theta^*),$$
where $J(\theta)$ is defined in (\ref{obj}) and
\begin{align}U(\mu,\theta)=\mathbb{E}_{b,r}\left[\sum_{i=1}^N\left\{r_{t,i}-m_{\mu,\theta}(b_{t,i})\right\}^2\pi_{\theta}(b_t,i)\right].\label{obj_mu}\end{align}
Under (\ref{critic}) and (\ref{actor}) which satisfy the compatibility condition, $\theta^*=\mathrm{argmax}_{\theta}J(\mu^*,\theta)$, where
$J(\mu,\theta)$ is redefined in (\ref{obj_theta}).

We assume that the arguments that achieve the maximum($\mathrm{argmax}$) and minimum($\mathrm{argmin}$) both exist in the parameter space that we consider. While the definition of $\theta^*$ is the same as the original definition,  we notice that the definition of $\mu^*$ now depends on the value of $\theta^*$. 

\subsection{Estimating Functions for $\mu^*$ and $\theta^*$}

The target parameters $\mu^*$ and $\theta^*$ are the values that jointly minimize (\ref{obj_mu}) with respect to $\mu$ and maximize (\ref{obj_theta}) with respect to $\theta$. To consistently estimate the parameters, we use as estimating functions the empirical versions of (\ref{obj_mu}) and (\ref{obj_theta}) that converge in probability to (\ref{obj_mu}) and (\ref{obj_theta}) respectively. Suppose we use the residual mean square (RMS) $\frac{1}{t}\sum_{\tau=1}^t\{r_{\tau,a_{\tau}}-m_{\mu, \theta}(b_{\tau,a_{\tau}})\}^2$ for (\ref{obj_mu}), which is computed on the context and reward pair of the chosen arms. Let $I_i(\tau)=I(a_{\tau}=i)$ be the binary indicator taking value $1$ if $a_{\tau}=i$ and $0$ otherwise. The expectation of the RMS is  
\begin{align*}
\mathbb{E}\left[\text{RMS}\right]&=\mathbb{E}\left[\frac{1}{t}\sum_{\tau=1}^t\sum_{i=1}^N\{r_{\tau,i}-m_{\mu, \theta}(b_{\tau,i})\}^2I_i(\tau)\right]\nonumber\\&=\mathbb{E}\left[\frac{1}{t}\sum_{\tau=1}^t\sum_{i=1}^N\{r_{\tau,i}-m_{\mu, \theta}(b_{\tau,i})\}^2\mathbb{E}[I_i(\tau)|\mathcal{F}_{\tau-1}]\right]\nonumber\\&=\mathbb{E}\left[\frac{1}{t}\sum_{\tau=1}^t\sum_{i=1}^N\{r_{\tau,i}-m_{\mu, \theta}(b_{\tau,i})\}^2\pi_{\hat{\theta}_{\tau-1}}(b_{\tau},i)\right],
\end{align*}
where $\mathcal{F}_{t-1}$ denotes a filtration at time $t$, the union of the history $\mathcal{H}_{t-1}$ of observations  up to time $t-1$ and the context $b_t$ at time $t$, i.e., $\mathcal{F}_{t-1}=\mathcal{H}_{t-1}\cup \{b_t\}$ where $\mathcal{H}_{t-1}=\bigcup_{\tau=1}^{t-1}\{b_{\tau}, a_{\tau}, r_{\tau, a_{\tau}}\}$. 
Due to Azuma-Hoeffding's inequality, the RMS converges in probability to $\mathbb{E}[\text{RMS}]$ for any $\mu$ and $\theta$. 
A gap with (\ref{obj_mu}) is caused by the udpate of $\hat{\theta}_{\tau}$ at each time point. To resolve this, we propose to minimize the following importance-weighted RMS instead,  
\begin{align}\hat{U}^t(\mu,\theta)&=\frac{1}{t}\sum_{\tau=1}^t\{r_{\tau, a_{\tau}}-m_{\mu, \theta}(b_{\tau, a_{\tau}})\}^2\frac{\pi_{\theta}(b_{\tau},a_{\tau})}{\pi_{\hat{\theta}_{\tau-1}}(b_{\tau},a_{\tau})}\\
&=\frac{1}{t}\sum_{\tau=1}^t\sum_{i=1}^N\{r_{\tau, i}-m_{\mu, \theta}(b_{\tau, i})\}^2\pi_{\theta}(b_{\tau},i)\frac{I_{i}(\tau)}{\pi_{\hat{\theta}_{\tau-1}}(b_{\tau},i)}\label{R4requestU}\end{align}
Since $\mathbb{E}[I_i(\tau)|\mathcal{F}_{\tau-1}]=\pi_{\hat{\theta}_{\tau-1}}(b_{\tau},i),$ the expectation of $\hat{U}^t(\mu,\theta)$ is exactly (\ref{obj_mu}), and $\hat{U}^t(\mu,\theta)$ converges in probability to (\ref{obj_mu}) for any $\mu$ and $\theta$.

We note here that if we had the guarantee that $\hat{\theta}_t$ converges in probability to $\theta^*,$ then the RMS would converge to (\ref{obj_mu}) as well. However, the convergence of $\hat{\theta}_t$ to $\theta^*$ is guaranteed only when the compatibility condition holds, which requires itself that the RMS converges to (\ref{obj_mu}).

The empirical version of (\ref{obj_theta}) is
\begin{align}\hat{J}^t(\mu,\theta)=\frac{1}{t}\sum_{\tau=1}^t\sum_{i=1}^Nb_{\tau,i}^T\mu~\pi_{\theta}(b_{\tau},i)-\lambda\theta^T\theta,\label{R4requestJ}\end{align}
and its expectation is exactly (\ref{obj_theta}). In the next section, we prove that the values of $\mu$ and $\theta$ that minimize $\hat{U}^t(\mu,\theta)$ with respect to $\mu$ and maximize $\hat{J}^t(\mu,\theta)$ with respect to $\theta$ converge in probability to $\mu^*$ and $\theta^*$ respectively.

\subsection{Computation}

\label{Propalg}
\begin{algorithm}[tb]
\label{alg2}
\caption{Actor-Improper Critic algorithm}
\begin{algorithmic}[1]
\STATE Set $\lambda>0$, $C>1$,  $\hat{\theta}_0=0_d$.
\FOR{$t=1,\cdots, T$}
\STATE {Pull arm $a_t$ according to probability $\left\{{\pi}_{\hat{\theta}_{t-1}}(b_t,i)\right\}_{i=1}^N$}  and get reward $r_{t,a_t}$.
\STATE {\bf Critic update}: $\hat{\mu}_t\leftarrow\underset{\mu: ||\mu||_2\leq C}{\mathrm{argmin}}\hat{U}^t(\mu,\hat{\theta}_{t-1})$  (see (\ref{R4requestU}))
\STATE {\bf Actor update}:  $\hat{\theta}_t\leftarrow\underset{\theta}{\mathrm{argmax}}~\hat{J}^t(\hat{\mu}_t,\theta)$  (see (\ref{R4requestJ})).
\ENDFOR
\end{algorithmic}
\end{algorithm}

The proposed algorithm with the new estimating functions is presented in Algorithm 2. {At each iteration of the algorithm, 
we find the value $\hat{\mu}_t$ which minimizes $\hat{U}^t(\mu, \theta)$ with $\theta$ replaced with $\hat{\theta}_{t-1}$ from the previous iteration. Then we find the value $\hat{\theta}_t$ which maximizes $\hat{J}^t(\mu, \theta)$ with $\mu$ replaced with $\hat{\mu}_t$. }
The inverse probability $1/\pi_{\hat{\theta}_{\tau-1}}(b_{\tau},i)$ can have large value and cause instability of the estimate $\hat{\mu}_t.$ To mitigate such instability, we solve $\hat{\mu}_t=\underset{\mu: ||\mu||_2\leq C}{\mathrm{argmin}}\hat{U}^t(\mu, \theta)$ for some positive constant $C$. We later show that if $C$ is set such that $\mu^*\in \{\mu: ||\mu||_2\leq C\},$
$\hat{\mu}_t$ and $\hat{\theta}_t$ converge in probability to $\mu^*$ and $\theta^*$ respectively. Without the constraint, $\hat{\mu}_t$ is just a weighted least-squares estimator with importance weights ${\pi_{\hat{\theta}_{t}}(b_{\tau},a_{\tau})}/{\pi_{\hat{\theta}_{\tau-1}}(b_{\tau},a_{\tau})}$'s. We later show that $\hat{\mu}_t$ with the constraint converges to the weighted least-squares estimator as time accumulates.

\subsection{Regret Bound}

The proposed algorithm (Algorithm 2) is robust to the misspecification of the critic model and converges to the optimal action selection policy. We define the regret with respect the optimal action selection policy as follows,
$$R(T)=\sum_{t=1}^T\sum_{i=1}^N\mathbb{E}[r_{t,i}|b_{t,i}]\left\{\pi_{\theta^*}(b_{t},i)-\pi_{\hat{\theta}_{t-1}}(b_{t},i)\right\}.$$
We can show that the proposed algorithm achieves a regret that is upper-bounded by $O(\sqrt{T})$ with high probability. This upper bound is of same order as the regret upper bound of Algorithm 1 which requires a restrictive assumption that the linear model correctly specifies the reward model. We provide the proofs in the Supplementary Material.

\subsection{Asymptotic Properties and Testing Procedure}

Statistical tests on the significance of the true parameter values ($\mu^*$ and $\theta^*$) can be conducted if the distribution of the estimates are known. In this section, we derive the asymptotic distribution of $\hat{\mu}_t$ and $\hat{\theta}_t$. We first state some necessary assumptions.

\begin{assump}\label{A1} The distribution of contexts variables is i.i.d. over time $t$, i.e., 
$$b_t=\{b_{t,1},\cdots, b_{t,N}\}\overset{i.i.d.}{\sim} P_b,$$
where $P_b$ is some distribution over $\mathbb{R}^{N\times d}$. {Also, the distribution of rewards $r_t=\{r_{t,1},\cdots,r_{t,N}\}$ is i.i.d. over time $t$.}
\end{assump}

\begin{assump}\label{A3}
Contexts and rewards are bounded. Without loss of generality, $||b_{t,i}||_2\leq 1$ and $|r_{t,i}|\leq 1$.
\end{assump}

\begin{assump}\label{A2}
The optimal policy $\theta^*$ is unique and $\mu^*$ is unique, and the joint equation $\left[\left\{\frac{\partial}{\partial \mu}U(\mu,\theta)\right\}^T, \left\{\frac{\partial}{\partial \theta}J(\mu,\theta)\right\}^T\right]=0_{2d}^T$ has unique solution at $[\mu^T,\theta^T]=[\mu^{*T},\theta^{*T}]$. Moreover, for a fixed value of $\mu,$ $J(\mu, \theta)$ has unique maximum at $\theta=\theta^*_{\mu}$. Also for a fixed value of $\theta,$ $U(\mu, \theta)$ has unique minimum at $\mu=\mu^*_{\theta}.$

\end{assump}

\begin{assump}\label{A4}
Let $\bar{b}_{\theta}(t)=\sum_{i=1}^N\pi_{\theta}(b_t,i)b_{t,i}$. The matrix $\mathbb{E}_{\theta}[(b_{t, a_t}-\bar{b}_{\theta}(t))(b_{t, a_t}-\bar{b}_{\theta}(t))^T]=\mathbb{E}\left[\sum_{i=1}^N\pi_{\theta}(b_t,i)(b_{t,i}-\bar{b}_{\theta}(t))(b_{t,i}-\bar{b}_{\theta}(t))^T\right]$ is positive definite for $\theta$ in a neighborhood of $\theta^*.$
\end{assump}

{Assumption \ref{A1} is standard in literature \citep{Langford07, Goldenshluger13, Bastani20} and is reasonable in many practical settings such as clinical trials where arms have a stationary distribution and do not depend on the past.} {The uniqueness of $\mu^*$ follows under mild conditions as it minimizes a convex function (\ref{obj_mu}) and because we can discard all the contextual features which do not differ by arms.} {The uniqueness of $\theta^*$ is a reasonable assumption as well since the penalty $-\lambda\theta^T\theta$ in (\ref{obj}) introduces a degree of convexity. Also due to this penalty $||\theta^*||_2$ is bounded, so the optimal policy itself is a policy that enforces a positive probability ($\gamma$) of uniform exploration. Therefore, we have $\mathbb{E}_{\theta^*}[(b_{t, a_t}-\bar{b}_{\theta^*}(t))(b_{t, a_t}-\bar{b}_{\theta^*}(t))^T]\succeq \gamma \mathbb{E}\left[\sum_{i=1}^N(b_{t,i}-\bar{b}_{\theta^*}(t))(b_{t,i}-\bar{b}_{\theta^*}(t))^T\right].$
Positive-definiteness of the right-hand side imposes variety in the arm features and  is also a standard assumption in the literature. (See \citet{Goldenshluger13} and \citet{Bastani20}.) Hence, Assumption 4 is reasonable as well.} 

We first show the following lemma which is crucial in deriving the consistency of the estimates.

\begin{lem}\label{bdd}
Under Assumptions \ref{A3}-\ref{A4}, the optimal policy parameter $\theta^*$ and $\mu^*$ lie in a compact set. Also,  the estimated parameters $\hat{\mu}_t$ and $\hat{\theta}_t$ lie in a compact set for all $t\in [T]$.
\end{lem}
\begin{proof}
We first show the boundedness of $\mu^*$. Since $\mu^*$ is the minimizer of $U(\mu,\theta^*)$, we have 
\begin{align*}\mu^*&=\left\{\mathbb{E}\left[\sum_{i=1}^N\pi_{\theta^*}(b_t,i)(b_{t,i}-\bar{b}_{\theta^*}(t))(b_{t,i}-\bar{b}_{\theta^*}(t))^T\right]\right\}^{-1}\times \mathbb{E}\left[\sum_{i=1}^N\pi_{\theta^*}(b_t,i)(b_{t,i}-\bar{b}_{\theta^*}(t))r_{t,i}\right].\end{align*} 
Due to Assumption \ref{A3}, we have $||\mu^*||_2\leq 1/\phi^2$ where $$\phi^2 =\lambda\left(\mathbb{E}\left[\sum_{i=1}^N\pi_{\theta^*}(b_t,i)(b_{t,i}-\bar{b}_{\theta^*}(t))(b_{t,i}-\bar{b}_{\theta^*}(t))^T\right]\right)$$
and $\lambda(\cdot)$ denotes the minimum eigenvalue. Due to Assumption \ref{A4} $\phi^2>0$, so $\mu^*$ lies in a compact set. Now, since $\theta^*$ maximizes $J(\mu^*,\theta)$, we have 
$J(\mu^*, 0_d)\leq J(\mu^*, \theta^*).$ Due to $||\mu^*||_2\leq 1/\phi^2$, Assumption \ref{A3},  and Cauchy-Schwarz inequality, we have $J(\mu^*, \theta^*)\leq \frac{1}{\phi^2}-\lambda\theta^{*T}\theta^*.$ Also, $J(\mu^*,0_d)\geq -\frac{1}{\phi^2}.$ Therefore, we have 
$-\frac{1}{\phi^2}\leq \frac{1}{\phi^2}-\lambda\theta^{*T}\theta^*$
which shows that $||\theta^*||_2\leq \sqrt{\frac{2}{\lambda\phi^2}}$. Due to line 4 of Algorithm 2, $||\hat{\mu}_t||_2\leq C$ so $\hat{\mu}_t$ clearly lies in a compact set, and analogously, we can show that $||\hat{\theta}_t||_2\leq \sqrt{\frac{2C}{\lambda\phi^2}}$. 
\end{proof}

We now prove in Theorem 1 the consistency of the estimates $\hat{\mu}_t$ and $\hat{\theta}_t.$

\begin{thm} {\bf Consistency}
Let $C^*=1/\phi^2$. Under assumptions \ref{A1}-\ref{A4}, if $C^*\leq C$, $(\hat{\mu}_t^T, \hat{\theta}_t^T)$ converges to $(\mu^{*T}, \theta^{*T})$ in probability.
\end{thm}
\begin{proof} We denote $\Omega=\{(\mu^T,\theta^T): ||\mu||_2\leq C, ||\theta||_2\leq 2\sqrt{2C/(\lambda\phi^2)}\}.$ Then $\Omega$ forms a compact set and includes $(\mu^{*T}, \theta^{*T})$. Since $\sum_{i=1}^Nb_{\tau,i}^T\mu~\pi_{\theta}(b_{\tau},i)$ is i.i.d. over time $\tau$ for fixed $\mu$ and $\theta$, we can apply Glivenko-Cantelli Theorem to $\hat{J}^t(\mu, \theta)$ and prove uniform convergence.  
\begin{align}    
    \underset{(\mu^T, \theta^T)\in\Omega}{\mathrm{sup}}\left|{\hat{J}^t(\mu,\theta)}-{{J}(\mu,\theta)}\right|\overset{P}{\longrightarrow} 0. \label{unifconv_theta}
\end{align}
Due to the term $I_i(\tau)/\pi_{\hat{\theta}_{\tau-1}}(b_{\tau},i)$, $\hat{U}^t(\mu,\theta)$ is not the mean of i.i.d. variables and requires additional steps to prove the uniform convergence. Define $$\tilde{U}^t(\mu,\theta)=\frac{1}{t}\sum_{\tau=1}^t\sum_{i=1}^N\{r_{\tau,i}-m_{\mu,\theta}(b_{\tau,i})\}^2\pi_{\theta}(b_{\tau},i).$$ Using martingale inequalities along with a covering argument on the space $\Omega$, we first show that $|\hat{U}^t(\mu,\theta)-\tilde{U}^t(\mu,\theta)|$ converges uniformly to 0 in probability. Then we apply Glivenko-Cantelli theorem to $\tilde{U}^t(\mu,\theta)$ to finally prove
\begin{align}
    \underset{(\mu^T, \theta^T)\in\Omega}{\mathrm{sup}}\left|{\hat{U}^t(\mu,\theta)}-{{U}(\mu,\theta)}\right|\overset{P}{\longrightarrow} 0, \label{unifconv_mu}\end{align} 
Since $(\hat{\mu}_t^T, \hat{\theta}_t^T)$ lies in $\Omega$ (Lemma 2), ${{U}(\mu,\theta)}^T$ and ${J}(\mu,\theta)^T$ are continuous on $\Omega$, and $(\mu^{*T}, \theta^{*T})$ is unique (Assumption \ref{A2}), we can apply Theorem 9.4 in \citet{Keener10} to show $(\hat{\mu}_t^T, \hat{\theta}_t^T)\overset{P}{\longrightarrow}(\mu^{*T}, \theta^{*T})$. Detailed proofs are presented in the Supplementary Material. 
\end{proof}

\begin{lem}\label{lem3} Suppose $C^*\leq C$. As $t\rightarrow \infty$, $\hat{\mu}_t$ converges in probability to the solution of $\hat{U}^t_{\mu}(\mu, \hat{\theta}_{t-1})=0$, where $\hat{U}^t_{\mu}(\mu,\theta)=\frac{\partial}{\partial \mu}\hat{U}^t(\mu,\theta)$ and $\hat{J}^t_\theta(\mu,\theta)=\frac{\partial}{\partial \theta}\hat{J}^t(\mu,\theta)$.

\begin{proof}
We just need to show that the solution $\tilde{\mu}_t$ of $\hat{U}^t_{\mu}(\mu, \hat{\theta}_{t-1})=0$ satisfies $P(\tilde{\mu}_t\leq C)\underset{t\rightarrow \infty}{\longrightarrow} 1,$ i.e.,  $P(\hat{\mu}_t=\tilde{\mu}_t)\rightarrow 1.$ Note that the solution of $\hat{U}^t_{\mu}(\mu, \hat{\theta}_{t-1})=0$ is a weighted least-squares estimator with weights $w_{\tau}={\pi_{\hat{\theta}_{t-1}}(b_{\tau},a_{\tau})}/{\pi_{\hat{\theta}_{\tau-1}}(b_{\tau},a_{\tau})}$, covariates $b_{\tau, a_{\tau}}-\bar{b}_{\hat{\theta}_{t-1}}(\tau)$, and outcomes $r_{\tau, a_{\tau}}$.  The lemma holds due to Assumption \ref{A3}, \ref{A4}, and the consistency of $\hat{\theta}_{t-1}$. Detailed proof can be found in the Supplementary Material.
\end{proof}
\end{lem}

\begin{thm} {\bf Asymptotic Normality}
Under assumptions \ref{A1}-\ref{A4}, if $C^*\leq C$, $\sqrt{t}\left(\left(\begin{array}{c}\hat{\mu}_t\\ \hat{\theta}_t\end{array}\right)-\left(\begin{array}{c}{\mu}^*\\ {\theta}^*\end{array}\right)\right)$ converges in distribution to a multivariate normal distribution with mean 0 and variance $\Psi=\Lambda^{-1}V^*\Lambda^{-1}$
where 
$$\Lambda=\left[\begin{array}{cc}U_{\mu\mu}^*& U_{\mu\theta}^*\\J_{\theta\mu}^*&J_{\theta\theta}^*\end{array}\right],$$ $$V^*=\underset{t\rightarrow\infty}{lim}\frac{1}{t}\sum_{\tau=1}^t\mathbb{E}\left[\left[\begin{array}{cc}u_{\mu}^{\tau *}u_{\mu}^{\tau *T}&u_{\mu}^{\tau *}j_{\theta}^{\tau *T}\\j_{\theta}^{\tau *}u_{\mu}^{\tau *T}&j_{\theta}^{\tau *}j_{\theta}^{\tau *T}\end{array}\right]\Bigg|\mathcal{F}_{\tau-1}\right],$$
\begin{align*}
    u_{\mu}^{\tau}(\mu,\theta)&=-\sum_{i=1}^N2\{r_{\tau,i}-m_{\mu, \theta}(b_{\tau,i})\}\dot{m}_{\mu, \theta}(b_{\tau,i})\times\frac{I_i(\tau)}{\pi_{\hat{\theta}_{\tau-1}}(b_{\tau},i)}\pi_{\theta}(b_{\tau},i),\\
    j_{\theta}^{\tau}(\mu, \theta)&=\sum_{i=1}^Nb_{\tau,i}^T\mu\dot{\pi}_{\theta}(b_{\tau},i)-2\lambda\theta,
\end{align*}
$U_{\mu\mu}$ and $U_{\mu\theta}$ are second order partial derivatives of $U$ with respect to $\mu$ twice and with respect to $\mu$ and $\theta$ respectively, the $J_{\theta\mu}$ and $J_{\theta\theta}$ are defined analogously, and the $U_{\mu\mu}^*, U_{\mu\theta}^*, J_{\theta\mu}^*, J_{\theta\theta}^*, u_{\mu}^{\tau *}, j_{\theta}^{\tau *}$ are the values of $U_{\mu\mu}, U_{\mu\theta}, J_{\theta\mu}, J_{\theta\theta}, u_{\mu}^{\tau}, j_{\theta}^{\tau}$ evaluated at the true value $(\mu^{*T}, \theta^{*T})$.
The asymptotic variance $\Psi$ can be estimated by replacing the expectation operation $\mathbb{E}(\cdot)$ with the empirical mean and plugging-in the estimates  $(\hat{\mu}_t^T, \hat{\theta}_t^T)$. Due to Assumption \ref{A1} and consistency of $(\hat{\mu}_t^T, \hat{\theta}_t^T)$, such plug-in type estimator is consistent for the asymptotic variance. 

\begin{proof}
Due to Lemma \ref{lem3} and linearization method, for sufficiently large $t$,
\begin{align*}
    \left[\begin{array}{c}{0_d}\\{0_d}\end{array}\right]&=\left[\begin{array}{c}{\hat{U}^t_{\mu}(\hat{\mu}_t,\hat{\theta}_{t})}\\{\hat{J}^t_{\theta}(\hat{\mu}_t,\hat{\theta}_t)}\end{array}\right]=\left[\begin{array}{c}{\hat{U}^t_{\mu}({\mu}^*,{\theta}^*)}\\{\hat{J}^t_{\theta}({\mu}^*,{\theta}^*)}\end{array}\right]+\left[\begin{array}{cc}{\hat{U}^t_{\mu\mu}(\tilde{\mu},\tilde{\theta})}&{\hat{U}^t_{\mu\theta}(\tilde{\mu},\tilde{\theta})}\\{\hat{J}^t_{\theta\mu}(\breve{\mu},\breve{\theta})}&{\hat{J}^t_{\theta\theta}(\breve{\mu},\breve{\theta})}\end{array}\right]\left(\begin{array}{c}\hat{\mu}_t-\mu^*\\ \hat{\theta}_t-\theta^*\end{array}\right)
\end{align*}
where $\tilde{\mu}=\alpha\hat{\mu}_t+(1-\alpha)\mu^*, \tilde{\theta}=\alpha\hat{\theta}_t+(1-\alpha)\theta^*, \breve{\mu}=\beta\hat{\mu}_t+(1-\beta)\mu^*$ and $\breve{\theta}=\beta\hat{\theta}_t+(1-\beta)\theta^*$ for some $0\leq \alpha\leq 1$ and $0\leq \beta\leq 1$.
Due to the consistency of $(\hat{\mu}^T, \hat{\theta}^T)$ and the Law of Large Numbers,
\begin{align*}\sqrt{t}\left(\begin{array}{c}\hat{\mu}_t-\mu^*\\ \hat{\theta}_t-\theta^*\end{array}\right)&=-\left\{\left[\begin{array}{cc}{{U}_{\mu\mu}^*}&{{U}_{\mu\theta}^*}\\{{J}_{\theta\mu}^*}&{{J}_{\theta\theta}^*}\end{array}\right]+o_P(1)\right\}^{-1}\times \sqrt{t}\left[\begin{array}{c}{\hat{U}^t_{\mu}({\mu}^*,{\theta}^*)}\\{\hat{J}^t_{\theta}({\mu}^*,{\theta}^*)}\end{array}\right]\end{align*}

Since the $\hat{J}^t_{\theta}(\mu^*,\theta^*)$ is the empirical mean of i.i.d. variables with mean 0, we can apply the Central Limit Theorem (CLT) to derive the asymptotic distribution. On the other hand, $\hat{U}^t_{\mu}(\mu^*,\theta^*)$ is the empirical mean of $u_{\mu}^{\tau}(\mu^*,\theta^*)$ which are not i.i.d. due to the term $I_i(\tau)/\pi_{\hat{\theta}_{\tau-1}}(b_{\tau},i)$. Instead, the $u_{\mu}^{\tau}(\mu^*,\theta^*)$'s form a martingale difference sequence. Hence, we can apply martingale CLT to  $\sqrt{t}\left[{\hat{U}^t_{\mu}({\mu}^*,{\theta}^*)^T}{\hat{J}^t_{\theta}({\mu}^*,{\theta}^*)^T}\right]$ in whole and show that this converges to a normal distribution with mean 0 and variance $V^*$.

\end{proof}

\end{thm}

Based on Theorem 2, a $Z$-test can be conducted for a $j$-th variable ($j=1,\cdots,d$) using the test statistic $Z=\hat{\theta}_{t,j}/\sqrt{\Psi_{d+j,d+j}/t}$. We reject the null hypothesis  $H_0: \theta_j=0$ with significance level $\alpha$ when $2(1-\Phi(|Z|))<\alpha$, where $\Phi(\cdot)$ is the cumulative distribution function of the standard normal distribution. 

\section{Experiments}

\begin{table}[t]
  \centering
  \begin{tabular}{cccccc}
    \toprule
        \multirow{2}{*}{Param.}&\multicolumn{2}{c}{$\varepsilon$-greedy}&\multirow{2}{*}{Param.}&\multirow{2}{*}{AC }&\multirow{2}{*}{Proposed}\\        \cmidrule(r){2-3}
        & $i=1$ & $i=2$& & &  \\
    \cmidrule(r){1-3}\cmidrule(r){4-6}
    $\mu^i_{(i-1)d+1}$& 0.91 &0.93&$\theta_1$ & 0.77 &0.99\\
    $\mu^i_{(i-1)d+2}$& 0.93 &0.94&$\theta_2$ & 0.64 &0.99\\
    $\mu^i_{(i-1)d+3}$& 0.96 &0.90&$\theta_3$ & 0.74 &0.99\\
    $\mu^i_{(i-1)d+4}$& 0.58 &0.59&$\theta_4$ & 0.07 &0.09\\
    \bottomrule
  \end{tabular}
  \caption{Rejection rates of $H_0$ for each parameter (Param.) by $\varepsilon$-greedy \citep{Chen20}, Actor-Critic \citep{Lei17}, and Proposed algorithm}
\end{table}

\begin{figure}
    \centering
    \includegraphics[width=\linewidth]{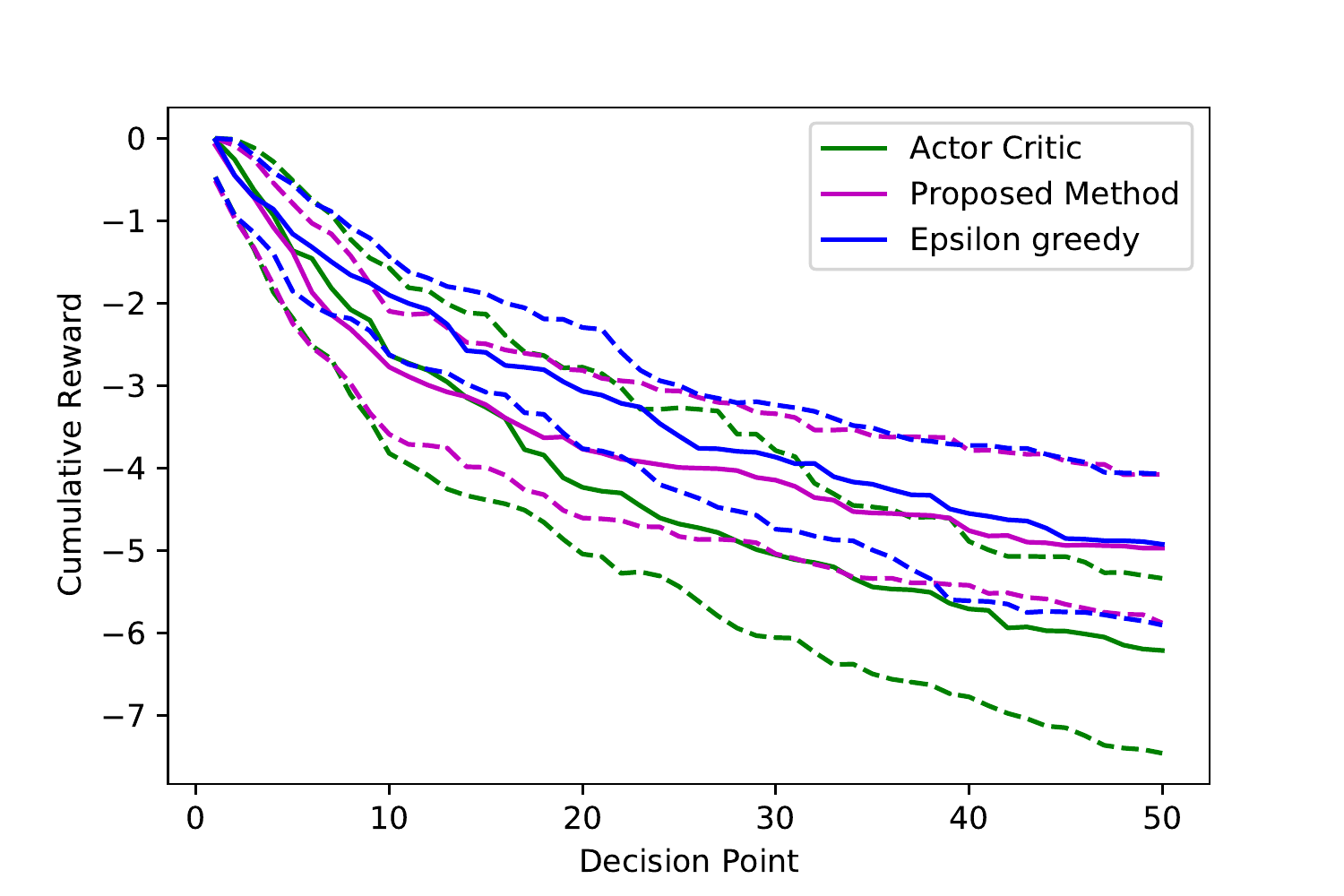}
    \caption{Median (solid lines) and first and third quartiles (dashed lines) of the cumulative rewards.}
    \label{fig:my_label}
\end{figure}

We conduct experiments to evaluate the performance of the Proposed algorithm under
a misspecified reward model. We set $N=2$ and $d=4$. 
We generate the context vectors $b_{t,i}$ from a multivariate normal distribution $\mathcal{N}(0_d, I^{d\times d})$ and truncate them to have $L_2$-norm 1. We generate the reward from a model nonlinear in $b_{t,i}$, $r_{t,i}=b_{t,i}^T\mu-\mathrm{max}(b_{t,1}^T\mu, b_{t,2}^T\mu)+\eta_{t,i}$ where $\mu=(-0.577, 0.577, 0.577, 0)^T$ and $\eta_{t,i}$ is generated from $\mathcal{N}(0, 0.01^2)$ independently over arms and time. {
To test the validity of the proposed testing procedure, we set $\mu_4$ to $0$ so that the corresponding variable does not affect the reward and hence, will not be useful in the policy. 

We implement the Proposed algorithm (Actor-Improper Critic) along with the original Actor-Critic algorithm \citep{Lei17} and the $\varepsilon$-greedy algorithm using weighted LSE \citep{Chen20}. When implementing the Proposed algorithm, we drop the restriction on $||\hat{\mu}_t||_2$ and compute the weighted least-squares estimator for simplicity. Since the objective function $J(\mu, \theta)$ is not convex with respect to $\theta$, we find the maximizer through a grid search followed by pattern search based on the Nelder-Mead method \citep{Nelder65} as suggested in \citet{Lei17}. When implementing the $\varepsilon$-greedy algorithm, we stack the context vectors into one vector $b_t^T=[b_{t,1}^T, \cdots, b_{t,N}^T]$ and set the working model for the reward of the $i$-th arm as $f_i(b_t)=b_t^T\mu^i$, where $\mu^i$ is a parameter of dimension $Nd$. We set the exploration parameter $\lambda$ in the AC and Proposed algorithms to $0.001$. In the $\varepsilon$-greedy algorithm, we use the value $\varepsilon=0.01$ which corresponds to the exploration probability guaranteed by $\lambda=0.001$ in the Proposed algorithm. We run the bandit algorithms until time horizon $T=50$ with 100 repetitions. 
}

For each algorithm, we count the number of times the null hypotheses $H_0:\theta_j=0$ (for AC and Proposed algorithms) or $H_0:\mu^i_{(i-1)d+j}=0$  (for $\varepsilon$-greedy) are rejected at time $T$ according to a $Z$-test  with significance level $\alpha=0.05$. We report the rejection rates of $H_0$ in Table 1. We observe that the AC algorithm \citep{Lei17} fails to reject the null hypotheses for the non-zero parameters $\theta_1, \theta_2$, and $\theta_3$ for at least 23\% of the experiments. On the other hand, the Proposed algorithm rejects the null hypotheses 99 times out of 100 times. As for the fourth parameter which has true value 0, both algorithms reject the null hypothesis with small probability. We note that the significance level $\alpha$ lies in the 95\% confidence interval $[0.047,0.133]$ computed from the formula $\left[\hat{p}-1.96\sqrt{{\alpha(1-\alpha)}/{n}}, \hat{p}+1.96\sqrt{{\alpha(1-\alpha)}/{n}}\right]$, where $\hat{p}=0.09$ is the rejection rate of $H_0$ by the Proposed algorithm and $n=100$ is the number of experiments. On the other hand, the $\varepsilon$-greedy algorithm rejects the null hypothesis for the fourth parameter with more than 50\% frequency. 
{We also note that the power of the tests for the $\varepsilon$-greedy algorithm with weighted LSE is lower than the Proposed algorithm. One possible reason for the low performance in testing is due to the variance induced by inverse probability weighting. The Proposed algorithm also involves computation of the inverse of probabilities, but it is used only in the ratio of probabilities $\pi_{\hat{\theta}_t}(b_{\tau},i)/\pi_{\hat{\theta}_{\tau-1}}(b_\tau,i)$ which converges to 1 as $\hat{\theta}_{\tau-1}$ converges.}
As for the cumulative rewards, we observe that the Proposed and the $\varepsilon$-greedy algorithm show comparable performance.

\section{Data Application}

 The Recovery Record Dataset contained  patients' adherence behaviors to their therapy for eating disorders (daily meal monitoring) and interactions with their linked clinicians on the app. Clinician communication is often viewed as a critical means to encourage adherence to monitoring, yet there is little guidance of when and how clinicians should communicate outside of office visits, and thus is  done on an ad-hoc and individual basis. The rewards (i.e. whether the patient adhered to daily monitoring) were observed for the actions chosen by the ad-hoc policies of clinicians (i.e. send a message or not).  A contextual bandit algorithm can allow clinicians to tailor their communications (arms) with patients who have preferences (contexts) to maximize adherence to therapy (rewards).
 
 We applied the offline policy evaluation method of \citet{Li11} to unbiasedly estimate the cumulative reward that we would obtain under the AC, Proposed, and $\varepsilon$-greedy algorithms. We repeated the evaluations on 30 bootstrap samples. {Table 2 shows the mean and standard deviations of the cumulative rewards at time $T=1000$.  We remark that although the Proposed and $\varepsilon$-greedy algorithms have comparable cumulative rewards, $\varepsilon$-greedy suffers higher variance. More details on the implementation and testing results can be found in the Supplementary Material. We show in the Supplementary material the rejection rates of $H_0$ for $\varepsilon$-greedy algorithms are closer to 0.5 as compared to the Proposed algorithm, which implies that $\varepsilon$-greedy may have lower power due to high variance.   }
 
 \begin{table}[t]
  \centering
  \begin{tabular}{cccc}
    \toprule
    &AC& Proposed & $\varepsilon$-greedy \\
    \midrule
    Mean & 5296.9 &5557.7&5577.3\\
    St.d. & 139.3 & 152.7 & 274.9 \\
    \bottomrule
  \end{tabular}
  \caption{Mean and standard deviations (St.d.) of cumulative rewards at $T=1000$ by $\varepsilon$-greedy \citep{Chen20}, Actor-Critic \citep{Lei17}, and Proposed algorithm for the Recovery Record Dataset}
\end{table}
 
\section{Conclusion}

The problem of testing the utility of variables collected by wearables or sensors represents a practical need in mHealth and is an inferential problem highlighted by Tewari and Murphy (2017). In this paper, we considered testing the utility of variables in the actor-critic bandit, but considered inference when the models used in the algorithms are not correctly specified.  This work demonstrates that a robust test can be constructed for the actor parameter.  Such work also illustrates that inferential procedures associated with the actor-critic bandit inherit problems such as model misspecification by virtue of the assumptions made in model-based testing; however, existing tools in the literature do not apply due to the unique structure of the objective function. This paper adds to the literature by developing a new statistical procedure to test the actor parameters in the actor-critic bandit even when the reward model is misspecified by the critic.

Strengths of this study include its contribution to model-based estimation in the context of contextual bandit algorithms, and the key property that these results are robust even when the assumptions underlying the critic fail to be correct. 

The ability to test variables may offer guidance in understanding what types of data (e.g. location or sensitive information) are useful.  Beyond the computational and performance related impact of this work, such knowledge could have societal impact in that it may discourage unnecessary data collection, thereby mitigating potential risks and threats to privacy. On the other hand, this method may also provide supporting evidence that certain types of data, perhaps either sensitive or costly variables, are in fact useful.  In such cases, the ethical tension between acquiring sensitive information or choosing not to acquire it at the cost of sub-optimal decision-making deserves disclosure and careful discussion with stakeholders involved in and affected by the algorithm.  This work provides a means to gauge the significance and assumptions made about the utility of data that would otherwise go untested, as it commonly does at present. 

While the focus of this paper is on testing, future work that focuses on implementation aspects of this procedure would be beneficial, such as addressing the question of when these tests should be performed in practice. The proposed method suggests that one feasible method to conduct testing after a large number of trials, but alternatives such as sequential or repeated hypothesis testing merit further attention.

\section*{Acknowledgments}

{Gi-Soo Kim is supported by the Institute of Information \& communications Technology Planning \& Evaluation (IITP) grant funded by the Korea government (MSIT) (No.2020-0-01336, Artificial Intelligence Graduate School Program(UNIST)) and the National Research Foundation of Korea (NRF) grant funded by the Korea government (MSIT, No.2021R1G1A100980111). Gi-Soo Kim is also partly funded by the 0000 Project Fund (Project Number 1.210079.01) of UNIST, South Korea.

Jane Paik Kim and Hyun-Joon Yang are supported by the National Institutes of Health Grant R01TR003505.}

\bibliographystyle{plainnat}  
\bibliography{Robust_tests_full}  

\newpage

\appendix

\numberwithin{equation}{section}
\setcounter{secnumdepth}{2}

\section*{\Large Supplementary Material}

In Section A, we derive the high-probability upper bound of the regret induced by the proposed algorithm. In Section B, we prove in detail Theorem 1 on the consistency of the parameter estimates. In Section C, we prove Lemma 3. Finally in Section D, we show the Data application results.

\section{Proof on Regret bound}

In Theorem 2, we have proved that $\sqrt{t}(\hat{\theta}_t-\theta^*)$ is asymptotically normal with distribution $\mathcal{N}(0_d, \Psi^{\theta})$ where $\Psi^{\theta}$ is the lower right $d\times d$ block matrix of $\Psi.$ We have also shown in Lemma 2 that $\hat{\theta}_t$ lies in a bounded set for all $t\in [T]$. Due to the two aforementioned facts, we have $$||\hat{\theta}_t-\theta^*||_2\leq O_{p}\left(\frac{1}{\sqrt{t}}\right).$$  
By decomposing the regret, we have,
\begin{align*}
R(T)&= \sum_{t=1}^T\sum_{i=1}^N\mathbb{E}[r_{t,i}|b_{t,i}]\left\{\pi_{\theta^*}(b_{t},i)-\pi_{\hat{\theta}_{t-1}}(b_{t},i)\right\}\\
&= \sum_{t=1}^T\sum_{i=1}^N\mathbb{E}[r_{t,i}|b_{t,i}]\dot{\pi}_{\check{\theta}_{t-1}}(b_t,i)^T\left\{{\theta^*}-{\hat{\theta}_{t-1}}\right\}
\end{align*}
for some $\check{\theta}_{t-1}\in\mathbb{R}^d$ which satisfies $||\check{\theta}_{t-1}-\theta^*||_2\leq ||\hat{\theta}_{t-1}-\theta^*||_2$ and $||\check{\theta}_{t-1}-\hat{\theta}_{t-1}||_2\leq ||\hat{\theta}_{t-1}-\theta^*||_2$, where $\dot{\pi}_{\check{\theta}_{t-1}}(b_t,i)$ is the derivative of $\pi_{\theta}(b_t,i)$ with respect to $\theta$ evaluated at $\theta=\check{\theta}_{t-1}$. Since $\check{\theta}_{t-1}$ is included in a bounded set, $||\dot{\pi}_{\check{\theta}_{t-1}}(b_t,i)||_2$ is bounded as well. Then by Cauchy-Schwarz inequality, we further have
\begin{align*}
    R(T)&\leq \sum_{t=1}^T\sum_{i=1}^NO_p\left(\frac{1}{\sqrt{t}}\right)=O_p(\sqrt{T}).
\end{align*}

\section{Proof of Theorem 1}

\subsection{Uniform convergence of $\hat{J}$ and  $\hat{U}$}

We denote $\Omega=\{(\mu^T,\theta^T): ||\mu||_2\leq C, ||\theta||_2\leq 2\sqrt{2C/(\lambda\phi^2)}\}.$ Then $\Omega$ forms a compact set and due to Lemma 2, $\Omega$ includes $(\mu^{*T}, \theta^{*T})$ and $(\hat{\mu}_t^T, \hat{\theta}_t^T)$ for every $t\in [T]$. 

Define $j(\mu,\theta,b)=\sum_{i=1}^Nb_i^T\mu\pi_{\theta}(b,i)-\lambda\theta^T\theta$. Then $\hat{J}^t(\mu,\theta)=\frac{1}{t}\sum_{\tau=1}^tj(\mu,\theta,b_{\tau})$ is the mean of i.i.d. random variables with expectation $J(\mu,\theta)=\mathbb{E}_b[j(\mu,\theta,b)]$. Since the class of random functions $\{j(\mu,\theta,b): (\mu^T,\theta^T)\in\Omega\}$ is Glivenko-Cantelli, it follows that 
$$\underset{(\mu^T, \theta^T)\in\Omega}{\mathrm{sup}}\left|{\hat{J}^t(\mu,\theta)}-{{J}(\mu,\theta)}\right|\overset{P}{\longrightarrow} 0. $$
On the other hand, $\hat{U}^t(\mu,\theta)$ is not the mean of i.i.d. variables due to the terms $I_i(\tau)/\pi_{\hat{\theta}_{\tau-1}}(b_{\tau},i)$, which requires additional steps to prove the uniform convergence. Define $$\tilde{U}^t(\mu,\theta)=\frac{1}{t}\sum_{\tau=1}^t\sum_{i=1}^N\{r_{\tau,i}-m_{\mu,\theta}(b_{\tau,i})\}^2\pi_{\theta}(b_{\tau},i).$$ Then using the same argument as for $\hat{J}^t(\mu,\theta)$, we can prove,
\begin{align}
    \underset{(\mu^T, \theta^T)\in\Omega}{\mathrm{sup}}\left|{\tilde{U}^t(\mu,\theta)}-{{U}(\mu,\theta)}\right|\overset{P}{\longrightarrow} 0.\label{firstu} 
\end{align}
Note that for fixed $\mu$ and $\theta$, $\hat{U}^t(\mu,\theta)-\tilde{U}^t(\mu,\theta)$ is the empirical mean of a martingale sequence. Also due to Assumption 2 and Lemma 2, the martingale sequence is bounded by a constant $B$. Then by Azuma-Hoeffding's inequality, we can show that for any $\varepsilon>0,$
$$\mathbb{P}\left(\left|\hat{U}^t(\mu,\theta)-\tilde{U}^t(\mu,\theta)\right|>\varepsilon\right)\leq 2\mathrm{exp}\left(-\frac{t\varepsilon^2}{2B^2}\right).$$ 
Let $\mathcal{N}_{\gamma}$ be a $\gamma$-covering number of the space $\Omega$ and $\Omega^{\gamma}$ be the corresponding cover. Then since $\Omega$ is a compact set in the $2d$-dimensional Euclidean space, we have $\mathcal{N}_{\gamma}=\left(\frac{G}{\gamma}\right)^{2d}$ where $G$ is a constant that depends on $C, \phi^2,$ and $\lambda.$ By union bound, we have 
\begin{align}\mathbb{P}\left(\left|\hat{U}^t(\mu,\theta)-\tilde{U}^t(\mu,\theta)\right|>\varepsilon\text{ for some } (\mu,\theta)\in\Omega^{\gamma}\right)&\leq \mathcal{N}_{\gamma}2\mathrm{exp}\left(-\frac{t\varepsilon^2}{2B^2}\right)\nonumber\\
&= 2\mathrm{exp}\left(-\frac{t\varepsilon^2}{2B^2}+2d\mathrm{log}\frac{G}{\gamma}\right)\label{ineq}\end{align} 
Now for any $(\mu,\theta) \in \Omega$, there exists $(\tilde{\mu},\tilde{\theta})\in\Omega^{\gamma}$ such that \begin{equation}\left|\hat{U}^t(\mu,\theta)-\tilde{U}^t(\mu,\theta)-\hat{U}^t(\tilde{\mu},\tilde{\theta})+\tilde{U}^t(\tilde{\mu},\tilde{\theta})\right|\leq (||\mu-\tilde{\mu}||_2+||\theta-\tilde{\theta}||_2)L\leq 2\gamma L\label{ineq2}\end{equation} for some constant $L$ where the first Lipschitz type inequality holds due to the continuity and boundedness of $\hat{U}^t(\cdot)$ and $\tilde{U}^t(\cdot)$.
Suppose we set $\gamma$ to depend on $t$ as $\gamma_t=O(1/\sqrt{t})$. Then from (\ref{ineq}) and (\ref{ineq2}), we have,
\begin{align*}
    \underset{(\mu^T, \theta^T)\in\Omega}{\mathrm{sup}}\left|{\hat{U}^t(\mu,\theta)}-{\tilde{U}^t(\mu,\theta)}\right|\leq 2\gamma_tL+\underset{(\tilde{\mu}^T, \tilde{\theta}^T)\in\Omega^{\gamma_t}}{\mathrm{sup}}\left|\hat{U}^t(\tilde{\mu},\tilde{\theta})-\tilde{U}^t(\tilde{\mu},\tilde{\theta})\right|\overset{P}{\longrightarrow} 0. 
\end{align*}
Combining this result with (\ref{firstu}), we can finally show
\begin{align*}
    \underset{(\mu^T, \theta^T)\in\Omega}{\mathrm{sup}}\left|{\hat{U}^t(\mu,\theta)}-{{U}(\mu,\theta)}\right|\overset{P}{\longrightarrow} 0.
\end{align*}


\subsection{Consistency of $\hat{\mu}_t$ and $\hat{\theta}_t$}

The remaining proof is organized as follows. We first denote the set of all possible values of $\mu$ and $\theta$ as $\Omega_{\mu}$ and $\Omega_{\theta}$ respectively. We first show that the solution $\hat{\theta}^t_\mu=\underset{\theta\in\Omega_{\theta}}{\mathrm{argmax}}\hat{J}^t(\mu,\theta)$ converges in probability to $\theta_{\mu}^*$ uniformly over $\Omega_{\mu}$, where  $\theta_{\mu}^*$ is defined in Assumption 3. We analogously show that $\hat{\mu}^t_{\theta}=\underset{\mu\in\Omega_{\mu}}{\mathrm{argmin}}\hat{U}^t(\mu,\theta)$ converges in probability to $\mu_{\theta}^*$ uniformly over $\Omega_{\theta}$. Finally, we show that $\hat{\mu}_t=\hat{\mu}^t_{\hat{\theta}_{t-1}}$ and $\hat{\theta}_t=\hat{\theta}^t_{\hat{\mu}_t}$ converge in probability to $\mu^*$ and $\theta^*$ respectively.

\subsubsection{Convergence of $\hat{\theta}^t_{\mu}$ and $\hat{\mu}^t_{\theta}$}

We apply Theorem 9.4 in \citet{Keener10} and follow the proof therein. Let $\Omega_{\theta}=\{\theta: ||\theta||_2\leq 2\sqrt{2C/(\lambda\phi^2)}\}$ and $\Omega_{\mu}=\{\mu: ||\mu||_2\leq C\}$. Fix a value of $\mu\in \Omega_{\mu}$. For sufficiently small $\varepsilon>0$, let $B_{\varepsilon}=\{\theta: ||\theta-\theta^*_{\mu}||_2\leq \varepsilon\}$. Denote $M=J(\mu,\theta^*_{\mu})=\underset{\theta\in\Omega_{\theta}}{\mathrm{sup}}J(\mu,\theta)$ and $M_{\varepsilon}=\underset{\theta\in\Omega_{\theta}\cap B_{\varepsilon}^C}{\mathrm{sup}}J(\mu,\theta)$. Since $\Omega_{\theta}\cap B_{\varepsilon}^C$ is compact as well and $J(\mu,\theta)$ is continuous on $\Omega$, $\exists \tilde{\theta}_{\mu}^{\varepsilon}\in \Omega_{\theta}\cap B_{\varepsilon}^C$ such that $J(\mu, \tilde{\theta}_{\mu}^{\varepsilon})=M_{\varepsilon}$. Due to Assumption 3, $M-M_{\varepsilon}=\Delta>0.$ 
Suppose that $$\underset{\theta\in\Omega_{\theta}}{\mathrm{sup}}\left|\hat{J}^t(\mu,\theta)-J(\mu,\theta)\right|<\frac{\Delta}{2}.$$
Then we have \begin{align*}
    \underset{\theta\in\Omega_{\theta}\cap B_{\varepsilon}^C}{\mathrm{sup}}\hat{J}^t(\mu,\theta)&\leq \underset{\theta\in\Omega_{\theta}\cap B_{\varepsilon}^C}{\mathrm{sup}}J(\mu,\theta)+\frac{\Delta}{2}=M_{\varepsilon}+\frac{\Delta}{2}=M-\frac{\Delta}{2}\\
    \underset{\theta\in\Omega_{\theta}}{\mathrm{sup}}\hat{J}^t(\mu,\theta)&\geq \hat{J}^t(\mu,\theta_{\mu}^*) > J(\mu,\theta_{\mu}^*)-\frac{\Delta}{2}=M-\frac{\Delta}{2}
\end{align*}
Hence, $\hat{\theta}^t_\mu\left(=\underset{\theta\in\Omega_{\theta}}{\mathrm{argmax}}\hat{J}^t(\mu,\theta)\right)$ should lie in $B_{\varepsilon.}$ This implies,
$$P\left(\underset{\theta\in\Omega_{\theta}}{\mathrm{sup}}\left|\hat{J}^t(\mu,\theta)-J(\mu,\theta)\right|<\frac{\Delta}{2}\right)\leq P\left(||\hat{\theta}^t_{\mu}-\theta_{\mu}^*||_2\leq \varepsilon\right).$$
The result holds for any $\mu\in\Omega_{\mu}$, i.e.,
$$P\left(\underset{\theta\in\Omega_{\theta}}{\mathrm{sup}}\left|\hat{J}^t(\mu,\theta)-J(\mu,\theta)\right|<\frac{\Delta}{2}\text{ for any }\mu\in\Omega_{\mu}\right)\leq P\left(||\hat{\theta}^t_{\mu}-\theta_{\mu}^*||_2\leq \varepsilon \text{ for any }\mu\in\Omega_{\mu}\right).$$
Therefore,
$$P\left(||\hat{\theta}^t_{\mu}-\theta_{\mu}^*||_2 > \varepsilon \text{ for some }\mu\in\Omega_{\mu}\right)\leq P\left(\underset{\theta\in\Omega_{\theta}}{\mathrm{sup}}\left|\hat{J}^t(\mu,\theta)-J(\mu,\theta)\right|\geq \frac{\Delta}{2} \text{ for some }\mu\in\Omega_{\mu}\right). $$
Since the right-hand side converges to 0, we have $\hat{\theta}^t_{\mu}\overset{P}{\longrightarrow}\theta_{\mu}^*$ uniformly over $\Omega_{\mu}$.

Analogously, we have $\hat{\mu}^t_{\theta}\left(=\underset{\mu\in\Omega_{\mu}}{\mathrm{argmin}}\hat{U}^t(\mu,\theta)\right)\overset{P}{\longrightarrow}\mu_{\theta}^*$ uniformly over $\Omega_{\theta}.$

\subsubsection{Convergence of ${\mu}^*_{\hat{\theta}_{t-1}}$ and ${\theta}^*_{\hat{\mu}_t}$}

Due to the above results, we have 
\begin{equation}\hat{\mu}_t=\hat{\mu}^t_{\hat{\theta}_{t-1}} \overset{P}{\longrightarrow} \mu^*_{\hat{\theta}_{t-1}}=\underset{\mu\in\Omega_{\mu}}{\mathrm{argmin}}~U\left(\mu,\hat{\theta}_{t-1}\right)\label{C1}\end{equation}
and
\begin{equation}\hat{\theta}_t=\hat{\theta}^t_{\hat{\mu}_t}\overset{P}{\longrightarrow} \theta^*_{\hat{\mu}_t}=\underset{\theta\in\Omega_{\theta}}{\mathrm{argmax}}~J\left(\hat{\mu}_t,\theta\right).\label{C2}\end{equation}
We will now prove that (i) $\tilde{\mu}_t=\mu^*_{\hat{\theta}_{t-1}}\overset{P}{\longrightarrow}\mu^*_{\theta^*}=\mu^*$ and (ii) $\tilde{\theta}_t=\theta^*_{\hat{\mu}_t}\overset{P}{\longrightarrow}\theta^*_{\mu^*}=\theta^*$. First, we show by contradiction that when either (i) or (ii) is true, then both should be true. We follow the proof technique of \citet{Lei17}. Suppose that (i) is true but (ii) is not. Then, 
\begin{equation}\exists \epsilon>0 ~\text{ s.t. }~ ||\tilde{\theta}_t -\theta^* ||>\epsilon ~\text{ for all }~t \label{not_maximum}\end{equation}
by taking a subsequence if necessary. For that subsequence, since $\tilde{\theta}_t$ is bounded, it converges to a constant $\tilde{\theta}$ by taking a subsequence if necessary. Due to (\ref{not_maximum}), we have \begin{equation}||\tilde{\theta}-\theta^*||>\epsilon\label{not_maximum2}.\end{equation}
Meanwhile, by definition of $\tilde{\theta}_t,$ we have
$$J(\hat{\mu}_t,\theta^*)\leq J(\hat{\mu}_t,\tilde{\theta}_t).$$
By continuity of $J()$, (\ref{C1}) and (i),  and convergence of $\tilde{\theta}_t$, we have
$$J(\mu^*,\theta^*)\leq J(\mu^*,\tilde{\theta}),$$
which is a contradictory result due to (\ref{not_maximum2}) and the uniqueness assumption (Assumption 3). Hence, we can conclude that when (i) is true, (ii) should be true. Similary, when (ii) is true, (i) should be true. We skip this proof as it is analogous.

We now show by contradiction that at least (i) or (ii) should be true. Combining this result with the precedent one gives the conclusion that both (i) and (ii) should be true. 

Suppose that (i) and (ii) are both wrong. Then,
\begin{equation}\exists \epsilon>0 ~\text{ s.t. }~ ||\tilde{\theta}_t -\theta^* ||>\epsilon~\text{ and  }~||\tilde{\mu}_t -\mu^* ||>\epsilon ~\text{ for all }~t \label{not_maximum_3}\end{equation}
by taking a subsequence if necessary. For that subsequence, since $\tilde{\mu}_t$ and $\tilde{\theta}_t$ are bounded, they converge to constants $\tilde{\mu}$ and $\tilde{\theta}$ respectively by taking a subsequence if necessary. Due to (\ref{not_maximum_3}), we have \begin{equation}||\tilde{\theta}-\theta^*||>\epsilon~\text{ and }~||\tilde{\mu}-\mu^*||>\epsilon\label{not_maximum4}.\end{equation}
Now suppose that $\tilde{\mu}\neq \mu^*_{\tilde{\theta}}.$ By definition of $\tilde{\mu}_t$, we have
$$U(\tilde{\mu}_t,\hat{\theta}_{t-1})\leq U({\mu}_{\tilde{\theta}}^*,\hat{\theta}_{t-1})$$
By continuity of $U()$, (\ref{C2}), and convergence $\tilde{\mu}_t$ and $\tilde{\theta}_t,$ we have
$$U(\tilde{\mu},\tilde{\theta})\leq U({\mu}_{\tilde{\theta}}^*, \tilde{\theta}),$$
which is a contradictory result due to uniqueness assumption (Assumption 3). Hence, we have $\tilde{\mu}=\mu_{\tilde{\theta}}^*$. Analogously, we have $\tilde{\theta}=\theta_{\tilde{\mu}}^*$. This implies that $\tilde{\mu}$ and $\tilde{\theta}$ are joint solvers of 
$$\frac{\partial}{\partial \mu}U(\mu,\theta)=0 ~\text{ and }~ \frac{\partial}{\partial \theta}J(\mu,\theta)=0,$$
which means $$\tilde{\mu}=\mu^*~\text{ and }~\tilde{\theta}=\theta^*,$$
which is a contradictory result due to (\ref{not_maximum4}). Therefore, at least (i) or (ii) should be true.

\section{Proof of Lemma 3}

Let $\tilde{\mu}_t$ be the solution of $\hat{U}_{\mu}^t(\mu, \hat{\theta}_{t-1})=0$. Then $\tilde{\mu}_t$ is a weighted least-squares estimator with weights $w_{\tau}={\pi_{\hat{\theta}_{t-1}}(b_{\tau},a_{\tau})}/{\pi_{\hat{\theta}_{\tau-1}}(b_{\tau},a_{\tau})}$, covariates $b_{\tau,a_{\tau}}-\bar{b}_{\hat{\theta}_{t-1}}(\tau)$, and outcomes $r_{\tau, a_{\tau}}$, i.e.,
$$\tilde{\mu}_t=\left\{\frac{1}{t}\sum_{\tau=1}^tw_{\tau}(b_{\tau, a_{\tau}}-\bar{b}_{\hat{\theta}_{t-1}}(\tau))(b_{\tau, a_{\tau}}-\bar{b}_{\hat{\theta}_{t-1}}(\tau))^T\right\}^{-1}\frac{1}{t}\sum_{\tau=1}^tw_{\tau}(b_{\tau, a_{\tau}}-\bar{b}_{\hat{\theta}_{t-1}}(\tau))r_{\tau, a_{\tau}}.$$
Due to consistency of $\hat{\theta}_t$ (Theorem 1) and continuity of $\pi_{\theta}()$ with respect to $\theta$, the denominator converges in probability to   $$\mathbb{E}\left[\sum_{i=1}^N\pi_{\theta^*}(b_t,i)(b_{t,i}-\bar{b}_{\theta^*}(t))(b_{t,i}-\bar{b}_{\theta^*}(t))^T\right].$$
Due to Assumption 4, the minimum eigenvalue (=$\phi^2$) of this matrix is positive. Also due to consistency of $\hat{\theta}_t$ (Theorem 1), continuity of $\pi_{\theta}()$ with respect to $\theta$, and Assumption 2, the nominator converges in probability to a vector with norm bounded by 1. Therefore, if $C\geq C^*=1/\phi^2,$  
 $P(\tilde{\mu}_t\leq C)\underset{t\rightarrow \infty}{\longrightarrow} 1,$ i.e.,  $P(\hat{\mu}_t=\tilde{\mu}_t)\underset{t\longrightarrow \infty}{\rightarrow} 1.$

\section{Data Application}

\subsubsection{Feature Construction} In order to inform the model in choosing an action, we engineered some features based on the interactions between clinicians and patients. These features include an indicator for whether the patient sent a message on a given day(\texttt{patient\_msg}), an indicator for whether the clinician asked a question(\texttt{question}), number of messages sent by the clinician(\texttt{num\_msg}), and the number of days since the last time the clinician sent a message(\texttt{last\_msg}). These variables were “lagged” by a day (i.e. based on interactions of the day prior to the action) to ensure they provided the context in which the clinician’s decision to take action was made rather than represent the interaction post facto. Demographic information was also incorporated through one-hot encoding the feature for the patient's gender(\texttt{gender}). For each unique clinician-patient pair, we derived the features each day from their initial interaction date to their last to generate all the data points.
 
\subsubsection{Probability of Action Selection by Logging policy} Probability of action selection for each data point was calculated using a logistic regression model fitted on the observed variables. The model took action-specific context variables as input and output the predicted action probability, which would serve as an estimate for an unbiased action selection probability conditioned on the observed variables. These probabilities were utilized to perform rejection sampling during offline evaluation so as the distribution of the context $b_t$ on which there is a match between the realized action in the dataset and the choice made by policies of interest ($\varepsilon$-greedy, Actor-Critic, and Proposed) is still i.i.d. over $t$. Details on the rejection sampling for unbiased offline evaluation can be found in \citet{Li11}.

\subsubsection{Hyperparameters-$\lambda$ and $\varepsilon$} 

We set the value of $\lambda$ for the Actor-Critic and Proposed algorithm as $\lambda=0.005$. The value of $\lambda$ controls the degree of exploration. We set the corresponding value of $\varepsilon$ for the $\varepsilon$-greedy algorithm according to the derivation in \citet{Lei17}. Define 
$$\theta^*_{\lambda}=\underset{\theta}{\mathrm{argmax}}~\mathbb{E}_{b,r}\left[\sum_{i=1}^Nr_{t,i}\pi_{\theta}(b_t,i)\right]-\lambda\theta^T\theta$$ 
and let $\beta_{\lambda}=\theta_{\lambda}^{*T}\theta^*_{\lambda}.$ Under assumption that the maximizer of $\mathbb{E}_{b,r}\left[\sum_{i=1}^Nr_{t,i}\pi_{\theta}(b_t,i)\right]$ is deterministic, i.e., $P\left(\pi_{\theta}(b_t,i)=1\right)>0$ for some $i$, we can show that $\theta^*_{\lambda}$ is the solution to the constrained problem,
$$\underset{\theta}{\mathrm{max}}~\mathbb{E}_{b,r}\left[\sum_{i=1}^Nr_{t,i}\pi_{\theta}(b_t,i)\right]~~\text{s.t.}~~\theta^T\theta\leq \beta_{\lambda}.$$
Here, $\theta^T\theta\leq \beta_{\lambda}$ means that for any $b\in\mathbb{R}^d$ with $||b||_2\leq 1$, $$|b^T\theta|\leq ||\theta||_2\leq \sqrt{\beta_{\lambda}}.$$
Hence, for the $N=2$ case, $$\frac{1}{1+\mathrm{exp}(\sqrt{\beta_{\lambda}})}\leq \pi_{\theta}(b_t,1)\leq \frac{\mathrm{exp}(\sqrt{\beta_{\lambda}})}{1+\mathrm{exp}(\sqrt{\beta_{\lambda}})}$$
We first run the Proposed algorithm and get and estimate $\widehat{\theta^*_{\lambda}}$ of $\theta^*_{\lambda}$. We then set $\widehat{\beta_{\lambda}}=\widehat{\theta^*_{\lambda}}^T\widehat{\theta^*_{\lambda}}$ and set $\varepsilon=\frac{1}{1+\mathrm{exp}(\sqrt{\widehat{\beta_{\lambda}}})}$. 

\subsubsection{Testing procedure and results}

In the Recovery Record Dataset, there is only one context vector $X_t\in\mathbb{R}^{d'}$ which contains variables that are not arm-specific. For Actor-Critic and Proposed algorithm, we construct $b_{t,1}$ and $b_{t,2}$ as 
\begin{align*}
    b_{t,1}^T&=[X_t^T, 0_{d'}^T]\in\mathbb{R}^d\\
    b_{t,2}^T&=[0_{d'}^T, X_t^T]\in\mathbb{R}^d
\end{align*}
where $0_{d'}$ is $d'$-dimensional zero vector and $d=2d'$. We accordingly set $\mu$ and $\theta$ as $d$-dimensional vectors. In this case, we have
\begin{align*}
    \pi_{\theta}(b_t,1)&=\frac{\mathrm{exp}(b_{t,1}^T\theta)}{\mathrm{exp}(b_{t,1}^T\theta)+\mathrm{exp}(b_{t,2}^T\theta)}=\frac{\mathrm{exp}((b_{t,1}-b_{t,2})^T\theta)}{1+\mathrm{exp}((b_{t,1}-b_{t,2})^T\theta)}
\end{align*}
where $$(b_{t,1}-b_{t,2})^T\theta=X_{t1}(\theta_1-\theta_{d'+1})+X_{t2}(\theta_2-\theta_{d'+2})+\cdots+X_{td'}(\theta_{d'}-\theta_{d'+d'}).$$ Therefore, to assess the utility of the variables, we should conduct a test on $\theta_j-\theta_{d'+j}$ for each $j\in[d']$. On the other hand, for the $\varepsilon$-greedy algorithm, we posit a linear model $f_i(X_t)=X_t^T\mu^i$ for each arm $i$ and conduct a test on $\mu^i_j$ for every $i\in[N]$ and $j\in[d']$. 

Table 3 shows the rejection rates of $H_0$ (under significance level 0.05) for either $\mu^i_j=0$ or $\theta_j-\theta_{d'+j}=0$ for every variable $j$ by $\varepsilon$-greedy\citep{Chen20}, Actor-Critic\citep{Lei17}, and Proposed algorithm. We observe that the Actor-Critic algorithm, which showed lowest reward, has rejection rate 0 for every variable. On the other hand, the $\varepsilon$-greedy and Proposed algorithm show a high rejection rate for \texttt{last\_msg}, indicating that this variable may be necessary for constructing the arm selection policy. We also observe that the Proposed algorithm has much lower rejection rate for \texttt{patient\_msg} and \texttt{num\_msg} as compared to the $\varepsilon$-greedy algorithm. One possible explanation is that the two variables are required to model the reward of each arm separately, but they are not necessary for explaining the difference in rewards between the arms.    

\begin{table}[t]
  \label{sample-table}
  \centering
  \begin{tabular}{ccccc}
    \toprule
        \multirow{2}{*}{Param.}&\multicolumn{2}{c}{$\varepsilon$-greedy}&\multirow{2}{*}{AC }&\multirow{2}{*}{Proposed}\\   
        \cmidrule(r){2-3}
        & $i=1$ & $i=2$& &\\   
    \cmidrule(r){1-5}
    \texttt{patient\_msg}& 0.666 & 0.533  &0  &0.2\\
    \texttt{num\_msg}& 0.6 & 0.633 & 0 &0.1\\
    \texttt{question}& 0.366 & 0.433 & 0 & 0.166\\
    \texttt{last\_msg}& 0.866 & 0.866& 0 & 0.9\\
    \texttt{gender\_F}& 0.6 & 0.666 &0 &0.5\\
    \texttt{gender\_M}& 0.366& 0.633 &0 &0.366\\
    \texttt{gender\_NA}& 0.333& 0.666 &0 &0.566\\
    \bottomrule
  \end{tabular}
  \caption{Rejection rates of $H_0$ for each parameter (Param.) by $\varepsilon$-greedy\citep{Chen20}, Actor-Critic \citep{Lei17}, and Proposed algorithm}
\end{table}

\end{document}